\documentclass[11pt]{article}
\usepackage{arxiv}
\usepackage{booktabs}
\usepackage{graphicx}
\usepackage{amsmath}
\usepackage{natbib}
\usepackage{url}

\newcommand{\modelid}{\texttt{google/\allowbreak diffusiongemma-\allowbreak 26B-\allowbreak A4B-\allowbreak it}}
\newcommand{\armodelid}{\texttt{google/\allowbreak gemma-\allowbreak 4-\allowbreak 26B-\allowbreak A4B-\allowbreak it}}
\newcommand{\acmethod}{\texttt{EntropyBoundSampler.\allowbreak accept\_canvas}}
\newcommand{\modelclass}{\texttt{DiffusionGemma\-For\-Block\-Diffusion}}

\runningtitle{Neither Parallel Nor Sequential}

\title{Neither Parallel Nor Sequential:\\ How DiffusionGemma Actually Commits Tokens}

\author{%
  Ali Asaria \\ Transformer Lab \and
  Tony Salomone \\ Transformer Lab \and
  Deep Gandhi\thanks{Corresponding author: \texttt{deep@lab.cloud}} \\ Transformer Lab
}

\date{}

\begin{document}
\maketitle

\begin{abstract}
Open diffusion language models are usually described as parallel, non-autoregressive
decoders, but the order in which a shipped checkpoint actually commits tokens is
rarely measured. We present an inference-only interpretability study of
\modelid{}, a masked discrete diffusion
mixture-of-experts model (25.2B total / 3.8B active parameters)
\ built on Gemma~4. We instrument its \texttt{transformers} denoising loop by
hooking \acmethod{}, recording, per accept-call,
which canvas positions are committed and their per-position entropy. Across a
686-prompt, six-regime probe suite,
\ in a strengthening run (5 seeds, 20 prompts/regime $=100$ traces/regime,
prompt-clustered bootstrap),
\ we find that DiffusionGemma exhibits a \emph{partial, granularity-dependent
left-to-right commit bias}, not clean block-autoregression. Token-level content commit
order is only moderate (tie-aware Kendall $\tau_b \approx 0.43$--$0.60$ for prose, code,
math and factual regimes)
\ and rises \emph{smoothly} with the analysis bin size: for R1, block-$\tau_b$ goes
$0.59 \to 0.64 \to 0.70 \to 0.79 \to 0.91$ as the bin grows $4 \to 8 \to 16 \to 32 \to 64$,
with no jump at 16,
\ so 16 is an analyst bin, not a privileged architectural block. The real token
order sits well \emph{below} a block-sequential control (token-$\tau_b \approx
0.94$--$0.96$ for R1/R2/R3/R5),
\ so there is genuine sub-block disorder, and commit batches are large: $\approx
13$--$26$ content tokens commit per accept-call,
\ with a large fraction of token pairs committing in the same call (same-call fraction
up to $0.72$),
\ so within-batch order is largely unresolved. The behaviour is regime-dependent:
structured JSON is approximately order-independent (token $\tau_b{=}-0.044$, 95\% CI
$[-0.086,-0.00]$).
\ Within a regime, commit confidence predicts correctness on math (GSM8K AUROC $0.749$,
prompt-clustered 95\% CI $[0.602,0.879]$)
\ but \emph{not} on factual recall (AUROC $0.471$, CI $[0.383,0.544]$).
\ Commitment is aggressive: a small late burst of accept-calls ($\ll$ the 48-step budget)
at entropy far below the 0.1 bound.
\ Commits are not strictly frozen (positions can re-mask), but the order conclusion is
robust on the strictly monotone subset.
\ Its task accuracy is comparable to the matched autoregressive sibling
Gemma-4~26B-A4B on the scorable regimes.
\ We also document the methodology this measurement demands, namely handling trailing
end-of-sequence (EOS) padding, Simpson's-paradox pooling, commit non-monotonicity,
block-size sensitivity, and commit-batch ties, as a contribution to how decoding
order should be measured.
\end{abstract}

\section{Introduction}

Diffusion language models (DLMs) generate text by iteratively denoising a canvas of
masked tokens, in principle committing many positions in parallel rather than strictly
left to right~\citep{2506.13759v5,2508.10875v2}. This framing motivates much of the
appeal of the paradigm: parallel commitment promises lower latency than
autoregressive (AR) decoding. Yet the \emph{order} in which a deployed checkpoint
actually finalizes tokens, and whether that order is sequential, parallel, or
something in between, is an empirical property of the trained model and its sampler,
not a given of the architecture. Recent work is split on what to expect: learned
decoding orders often trend left-to-right~\citep{2512.21446v2,2506.00413v2}, while
structured, out-of-order ``anchor-first'' generation is also documented
\citep{2602.02112v2,2601.15593v2}.

We study this question directly for a single, public, production
checkpoint:\footnote{The checkpoint we study is \modelid{}~\citep{diffusiongemma2026}, released open-weights under Apache-2.0.}
\ a masked discrete-diffusion mixture-of-experts model (25.2B total parameters, 3.8B
active via 8-of-128 expert routing)
\ built on Gemma~4. Crucially, this is an \emph{inference-only} study: we do not train
or fine-tune anything. Instead we instrument the model's own \texttt{transformers}
denoising loop, hooking its \acmethod{} method to
observe, at each accept-call, exactly which canvas positions are committed and with
what per-position entropy. From these traces we measure the decoding order, its
dependence on task regime, its granularity (bin-size) structure, whether commit confidence
is predictive of correctness, and an accuracy contrast against the matched AR
sibling Gemma-4~26B-A4B~\citep{gemma4}.

Our central finding is that DiffusionGemma exhibits a \emph{partial, granularity-dependent
left-to-right commit bias}, moderate at the token level and rising \emph{smoothly}
toward sequential at coarser analysis bins, rather than clean block-autoregression.
Three facts distinguish this partial bias from a clean block-autoregressive reading.
First, block-$\tau_b$ rises \emph{smoothly} with bin size, with no jump at 16, so 16 is an
analyst bin, not the model's architectural block (\S\ref{sec:results-order}).
\ Second, the real token order sits well below a pure block-sequential control, so there
is genuine sub-block disorder.
\ Third, commit batches are large ($\approx 13$--$26$ tokens per accept-call) with many
token pairs tied in the same call, so within-batch order is largely unresolved.
\ The bias is regime-dependent: structured JSON output is approximately
order-independent rather than strongly non-sequential. (An earlier two-seed run had read
this structure as ``block-autoregressive'' with a privileged 16-token block; the
strengthening run of 5 seeds with a prompt-clustered bootstrap revises that to the
partial bias reported here.)

\paragraph{Contributions.} We make the following falsifiable claims, each tied to a
section:
\begin{enumerate}
  \item \textbf{A partial, granularity-dependent left-to-right commit bias
  (\S\ref{sec:results-order}).} Token-level content commit order is only moderate
  (tie-aware Kendall $\tau_b \approx 0.43$--$0.60$ across prose, code, math and factual
  regimes), quantified with prompt-clustered 95\% CIs over 5 seeds, and rises
  \emph{smoothly} with the analysis bin size (e.g.\ R1: $0.59 \to 0.91$ as the bin grows
  $4 \to 64$, no jump at 16). The bias is real but partial; 16 is an analyst bin, not a
  privileged block.
  \item \textbf{Not cleanly block-sequential; large commit batches
  (\S\ref{sec:results-order}).} The real token-$\tau_b$ sits well below a pure
  block-sequential control ($\approx 0.94$--$0.96$), so there is genuine sub-block
  disorder; and $\approx 13$--$26$ content tokens commit per accept-call with a large
  same-call (tied) fraction, so within-batch order is largely unresolved and token order
  is measured at accept-call resolution.
  \item \textbf{Regime dependence: JSON is approximately order-independent
  (\S\ref{sec:results-order}).} Structured JSON has token $\tau_b{=}-0.044$ (95\% CI
  $[-0.086,-0.00]$), distinct from the moderate positive bias of the other regimes.
  \item \textbf{Regime-specific confidence--correctness discrimination
  (\S\ref{sec:results-calib}).} Commit confidence (negative entropy) predicts
  correctness on math (GSM8K AUROC $0.749$, prompt-clustered 95\% CI $[0.602,0.879]$;
  reliability monotone by entropy tertile) but not on factual recall (AUROC $0.471$, CI
  $[0.383,0.544]$). The signal is regime-specific, and cross-regime pooling is a
  Simpson's-paradox pitfall.
  \item \textbf{Aggressive entropy-bounded early stopping
  (\S\ref{sec:results-calib}).} Generations finish in a small burst of accept-calls, far
  below the 48-step budget, at entropy far below the 0.1 bound.
  \item \textbf{Accuracy comparable to the AR sibling
  (\S\ref{sec:results-ar}).} On the scorable regimes, DiffusionGemma's task accuracy is
  comparable to Gemma-4~26B-A4B; we did not power the comparison for an equivalence test
  (see \S\ref{sec:threats}).
\end{enumerate}
We also contribute the methodology for measuring decoding order honestly
(\S\ref{sec:threats}): the EOS-padding artifact, the Simpson's-paradox pooling pitfall,
commit non-monotonicity, block-size sensitivity, and commit-batch ties, caveats we
expect to recur in commit-order studies.

\section{Related Work}
\label{sec:related}

\paragraph{Decoding order in diffusion LMs.} The metric we adopt, Kendall's $\tau$
between a token's finalization step and its left-to-right index, is exactly the
``anchor-first parallelism'' lens of \citet{2601.15593v2}, who quantify generation
order in masked DLMs. Decoding order is increasingly treated as a controllable knob:
\citet{2601.20339v2} jointly search over generation order and token space, and
\citet{2602.02112v2} unify masked diffusion under various generation orders, tracking
which kinds of tokens commit early versus late. \citet{2512.21446v2} report that a
\emph{learned} order is predominantly left-to-right with code-specific structural
anchors, and \citet{2506.00413v2} show that a fixed left-to-right order with a
confidence threshold (adaptive parallel decoding) can match or beat
entropy/confidence/random orderings. Our work differs in target and method: rather
than proposing a decoding strategy, we \emph{measure} the order a shipped checkpoint
already produces, and we characterize it as a partial, granularity-dependent
left-to-right bias rather than clean block-autoregression.

\paragraph{Confidence- and entropy-guided commitment.} A second thread asks which
internal signal should drive unmasking. \citet{2603.22248v1} give a provable-efficiency
account of confidence-based decoding; \citet{2602.00250v2} steer unmasking by the
entropy of the \emph{future} masked set; and \citet{2511.05664v2} find that raw
confidence alone unmasks wrong tokens whereas step-to-step KL stability tracks
correctness. \citet{2512.21336v1} introduce a per-step ``denoising entropy'' as an
internal uncertainty signal correlated with quality. DiffusionGemma's shipped
entropy-bounded sampler is one concrete point in this design space; we treat its own
commit-entropy as the signal under test.

\paragraph{Calibration.} Whether model confidence reflects correctness is a calibration
question. \citet{2605.23909v1} study confidence calibration in LLMs with
difficulty-conditioned reliability, and \citet{2601.23096v2} calibrate confidence
during preference optimization. We import the warning that confidence need not track
correctness, and we probe it directly, finding that within-regime commit confidence
is \emph{predictive of} correctness on math (AUROC with a clustered CI excluding chance,
plus a monotone reliability-by-entropy-tertile curve) but \emph{null} on factual recall,
and that cross-regime pooling is a Simpson's-paradox pitfall. We report AUROC
(discrimination) and a tertile reliability curve, not a full ECE.

\paragraph{Surveys and scope.} Recent surveys~\citep{2506.13759v5,2508.10875v2} map the
DLM landscape. Several strong method papers train models from scratch~\citep{2506.13579v2,2604.26985v1,2605.06885v1}; we borrow their metrics and framings,
not their training, and stay strictly inference-only.

\section{Method}
\label{sec:method}

\paragraph{Subject model and decode configuration.} The subject is the public
checkpoint \modelid{}
\ (\modelclass{}, \texttt{model\_type diffusion\_gemma}, in
the \texttt{transformers} library, no \texttt{trust\_remote\_code}). It is a masked
discrete-diffusion mixture-of-experts model, 25.2B total / 3.8B active parameters
(8 of 128 experts),
\ built on Gemma~4. The ``26B-A4B'' in the vendor's name is a rounded marketing label
(${\approx}26$B total, ${\approx}4$B active); we use the parameter counts we actually
tallied (25.2B / 3.8B) in all technical statements. Its shipped decode configuration uses a 256-token canvas and up to
48 denoising steps,
\ with an \texttt{EntropyBoundSampler} whose \texttt{entropy\_bound} attribute is 0.1.
\ All runs use \texttt{transformers} 5.11.0, \texttt{torch} 2.9.0+cu128, on a single
NVIDIA H100 80GB.

\paragraph{Instrumentation: the \texttt{accept\_canvas} hook.} The model's sampler
exposes
\texttt{accept\_canvas(current\_canvas, denoiser\_canvas, logits, cur\_step)}, where
\texttt{logits} has shape $[1,256,262144]$ (vocabulary 262144).
\ This method is the model's commit mechanism: on each call it decides which masked
canvas positions to finalize. We wrap it with a forward hook that records, for every
position $p$, the \emph{first} accept-call index that commits $p$ (we write this
$\mathrm{commit}(p)$), together with the per-position entropy of the committing
logits (the entropy-at-commit). No model weights or sampling decisions are modified;
the hook is purely observational. Two consequences of this resolution matter for
interpretation. First, commit order is observed at \emph{accept-call} resolution: there
are only $\approx 3$--$17$ accept-calls per generation (\S\ref{sec:results-order}),
\ so many content positions commit on the same call and their within-call order is
unresolved; ``$\mathrm{commit}(p)$'' is therefore an accept-call index, not a
denoising-step or wall-clock order. Second, we record only the first accept-call that
commits a position; we do not verify that a committed position remains frozen on
subsequent calls, so ``commit order'' is precisely first-acceptance order
(\S\ref{sec:threats}).

\paragraph{Entropy-at-commit.} The entropy-at-commit of a position is the Shannon
entropy (in nats) of the committing accept-call's logits at that position, taken over
the full $262{,}144$-token vocabulary. It is reported in the same unit as the sampler's
\texttt{entropy\_bound} attribute (also nats), so the two are directly comparable.

\paragraph{Content positions.} Raw canvases are padded; trailing end-of-sequence (EOS)
positions are committed early as a degenerate side effect, which dominates a naive
order metric (see \S\ref{sec:threats}). We therefore restrict all order and
calibration statistics to \emph{content} positions: a position is content iff its final
token is not in \texttt{all\_special\_ids} $\cup \{0,1,2,\langle\text{end\_of\_turn}\rangle\}$.

\paragraph{Order metrics.} Our primary order statistic is the content-only tie-aware
Kendall rank correlation $\tau_b\big(\mathrm{commit}(p),\, p\big)$ between a content
token's commit-call index and its left-to-right canvas position. We use the
tie-corrected $\tau_b$ variant deliberately: because many positions share an accept-call
index (above), the commit ranks are heavily tied, and pairs that commit on the same
accept-call are \emph{simultaneous}, not strictly out of order. A value near $+1$
indicates strict left-to-right commitment; $0$ indicates order-independence; negative
values indicate a right-to-left tendency. To characterize granularity, we compute a
\emph{block-$\tau_b$} over a \emph{sweep} of bin sizes $B \in \{4,8,16,32,64\}$:
positions are grouped into contiguous $B$-token bins, each bin is assigned its earliest
(minimum) commit-call, and $\tau_b$ is computed between bin commit order and bin
position. The sweep replaces the earlier single 16-token bin: \textbf{16 is an
analyst-chosen analysis granularity, not the model's architectural block size}, and the
substantive claim is how $\tau_b$ varies with $B$ (\S\ref{sec:results-order}).

\paragraph{Block-sequential control.} To calibrate the $\tau_b$ scale, we compare the
real token order against a synthetic \emph{block-sequential} control in which each
position's commit time is set to $\mathrm{position}//16$ (a pure 16-block left-to-right
process, within-block tied). The control's token-$\tau_b$ bounds what clean block
autoregression would produce; the gap between it and the real $\tau_b$ measures genuine
sub-block disorder.

\paragraph{Commit batches and ties.} Because many positions commit on the same
accept-call, we also report the mean number of content tokens committed per accept-call
and the fraction of content token pairs that commit in the \emph{same} call (the
same-call/tied fraction). Large batches and a high tied fraction mean within-batch order
is unresolved and token-$\tau_b$ is measured at accept-call resolution. We additionally
report an out-of-order rate (fraction of content token pairs whose commit order disagrees
with their positional order); this is a monotone restatement of $\tau_b$ rather than an
independent statistic, and the reported out-of-order pairs include same-call ties.

\paragraph{Resampling and seeds.} All headline statistics come from a strengthening run
of 5 seeds ($\{11,22,33,44,55\}$) $\times$ 20 prompts per regime ($=100$ traces per
regime, 600 generations total).
\ Confidence intervals are 95\% \emph{prompt-clustered} bootstrap intervals (we resample
the 20 prompts, preserving within-prompt cross-seed correlation). We report the SD of the
mean-$\tau$ across the 5 seeds as a stability check (\S\ref{sec:results-order}),
\ replacing the earlier two-seed setup.

\paragraph{First-acceptance and freezing.} ``$\mathrm{commit}(p)$'' is the \emph{first}
accept-call that commits $p$. Commits are not strictly monotone: across the 600
generations we observe 4524 un-accept events (a previously committed position returns to
masked; mean $7.5$/gen, median $2$; only $220/600$ generations are fully monotone).
\ We therefore define commit order as first-acceptance order, and we check robustness by
recomputing $\tau_b$ on the strictly monotone subset of generations
(\S\ref{sec:results-order}).

\paragraph{Entropy and confidence--correctness analysis.} For each committed content
position we record the entropy-at-commit (nats). For the regimes with a correctness
label we form a binary correct/incorrect target per generation, aggregate each
generation's commit entropies to a per-generation confidence summary, and compute the
AUROC (area under the receiver-operating-characteristic curve) of
$-\text{entropy} \rightarrow \text{correct}$ with a \emph{prompt-clustered} 95\%
bootstrap CI, plus a reliability curve binning generations by entropy tertile
(low/medium/high) and reporting accuracy in each. We emphasize that AUROC quantifies
\emph{discrimination} (how well confidence separates correct from incorrect), not full
\emph{calibration}: we report the tertile reliability curve but no expected calibration
error (ECE). The signal is informative only where errors occur (\S\ref{sec:threats}). We
frame the results as commit-confidence being \emph{predictive of} correctness where the
clustered CI excludes chance, and \emph{null} where it does not. We deliberately analyze
this signal \emph{within} each regime; \S\ref{sec:threats} explains why pooling across
regimes is invalid.

\paragraph{Compute.} The complete study consumed approximately 0.9 H100-hours.

\section{Experimental Setup}
\label{sec:setup}

\paragraph{Probe suite.} We evaluate on a 686-prompt probe suite spanning six
regimes:
\ R1 math (GSM8K, 200 prompts), R2 code completion (HumanEval, 164), R3 code synthesis
(MBPP, 200), R4 short factual recall (52), R5 open-ended instructions (40), and R6
constrained JSON / key-value output (30). R1--R3 are sampled from the public Hugging
Face datasets-server (benchmark test splits of size 1319 / 164 / 500 respectively);
\ R4--R6 are authored from committed seed tables. The suite is built deterministically
(seed 20260611) and pinned by content hash (\texttt{probe\_suite.jsonl},
sha256 \texttt{20be4196}{\ldots}\texttt{40e1f}).

\paragraph{Held-out confirmatory protocol and strengthening run.} To avoid analysis
overfitting, the suite is split per-regime by content-hashed id into an exploratory half
(used to freeze all metric and threshold choices) and a confirmatory half (used only for
the reported numbers); no prompt straddles the split. The headline statistics below come
from a \emph{strengthening run} on held-out prompts: 5 seeds ($\{11,22,33,44,55\}$)
$\times$ 20 prompts per regime, i.e.\ $100$ traces per regime and $600$ generations
total.
\ Decoding uses the shipped configuration throughout (canvas 256, up to 48 steps,
entropy-bounded sampling). Confidence intervals are \emph{prompt-clustered} 95\% bootstrap
intervals: we resample the 20 prompts (with their 5 per-prompt seeds together), so the
intervals respect within-prompt cross-seed correlation. The SD of the mean-$\tau$ across
the 5 seeds is reported as a seed-stability check (\S\ref{sec:results-order}). This run
supersedes the earlier two-seed, non-clustered confirmatory numbers for all order and
calibration statistics.

\paragraph{Scope of correctness scoring.} Correctness is scored for R1 (GSM8K exact
match), R4 (factual exact match), and R6 (JSON validity). R2/R3 are not executed
against test cases (no code-execution harness), and R5 is open-ended with no reference,
so these regimes contribute to order analysis only.

\section{Results}

\subsection{Decoding order: a partial, granularity-dependent left-to-right bias}
\label{sec:results-order}

Table~\ref{tab:order} reports the per-regime order statistics from the strengthening
run. Token-level content commit order is \emph{moderately} left-to-right for prose,
code, math and factual regimes: tie-aware Kendall $\tau_b$ ranges from $0.430$ (R2) to
$0.604$ (R3),
\ all with prompt-clustered 95\% CIs above zero but far below the $+1$ of strict
autoregression. Because commit is observed at accept-call resolution and ranks are
heavily tied (\S\ref{sec:method}), these values should be read as an accept-call-level
order tendency, not a strict within-step ordering. Seed variance is small (SD of mean-$\tau$
across the 5 seeds is $0.014$--$0.050$),
\ so the values are stable. The confirmatory out-of-order rate is $0.086$--$0.208$,
\ a monotone restatement of $\tau_b$ rather than independent corroboration.

\begin{figure}[t]
\centering
\includegraphics[width=0.9\linewidth]{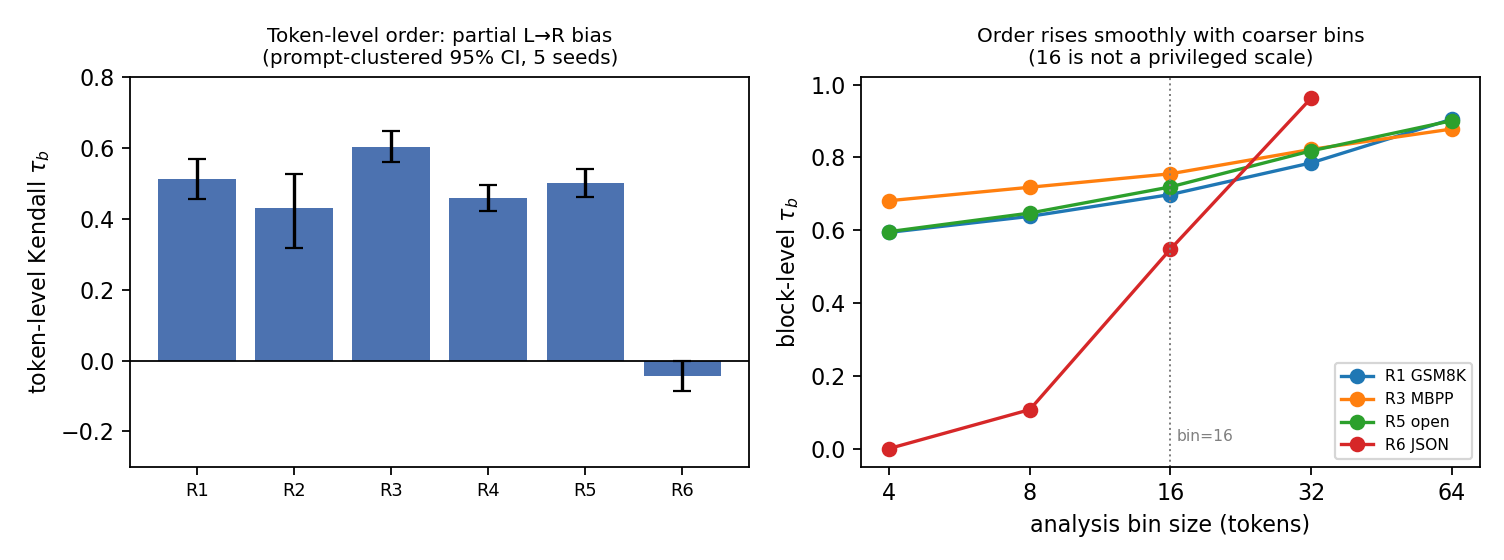}
\caption{Decoding order in the strengthening run. \emph{Left:} per-regime token-level
tie-aware Kendall $\tau_b$ with prompt-clustered 95\% CIs: a moderate left-to-right
bias for prose/code/math/factual, well below the $\tau_b=1$ of strict left-to-right
decoding, and approximately zero for structured JSON (R6).
\ \emph{Right:} block-$\tau_b$ versus analysis bin size $B \in \{4,8,16,32,64\}$. It
rises \emph{smoothly and monotonically} with $B$ (e.g.\ R1 $0.59 \to 0.91$), with no jump
at 16, so 16 is an analyst bin, not a privileged block.}
\label{fig:order}
\end{figure}

\paragraph{16 is not special: block-$\tau_b$ rises smoothly with bin size.} Grouping
positions into $B$-token bins and sweeping $B \in \{4,8,16,32,64\}$, block-$\tau_b$ rises
\emph{smoothly and monotonically} with $B$, with \emph{no jump at 16}. For R1 it goes
$0.59 \to 0.64 \to 0.70 \to 0.79 \to 0.91$ as $B$ grows $4 \to 8 \to 16 \to 32 \to 64$;%
[E37]
\ the other regimes show a similar monotone rise (Figure~\ref{fig:order}, right). Coarser
bins look more sequential simply because they hide more within-bin disorder; there is no
privileged architectural block at 16. The decoding order is therefore best described as a
\emph{partial, granularity-dependent} left-to-right bias, moderate at the token level,
rising smoothly toward sequential at coarser bins, not as block-autoregression with a
special block size.

\paragraph{Not cleanly block-sequential.} A pure block-sequential control
($\mathrm{commit}:=\mathrm{position}//16$, within-block tied) yields token-$\tau_b
\approx 0.94$--$0.96$ for R1/R2/R3/R5.
\ The real token-$\tau_b$ (Table~\ref{tab:order}) is only $0.43$--$0.60$, well below
the control, so the model is \emph{not} cleanly block-autoregressive: there is genuine
sub-block disorder. The gap between real and control $\tau_b$ is the quantitative size of
that disorder.

\paragraph{Large commit batches; within-batch order unresolved.} Commit batches are
large: $\approx 13$--$26$ content tokens commit per accept-call (R1 $25.5$, R2 $23.0$,
R3 $19.8$, R5 $12.6$; R4 $5.8$, R6 $8.5$),
\ and a large fraction of content token pairs commit in the \emph{same} call (same-call
fraction R1 $0.28$, R2 $0.49$, R3 $0.19$, R4 $0.50$, R5 $0.11$, R6 $0.72$).
\ Within-batch order is thus largely unresolved, and token-$\tau_b$ is measured at
accept-call resolution.

\begin{table}[t]
  \centering
  \caption{Per-regime decoding order on the held-out strengthening run (5 seeds
  $\times$ 20 prompts $=100$ traces per regime). Token $\tau_b$ is the content-only
  tie-aware Kendall correlation between commit-call index and left-to-right position
  (prompt-clustered 95\% bootstrap CI in brackets). ``Block-seq.\ control'' is the
  token-$\tau_b$ of a pure $\mathrm{position}//16$ process and bounds clean block
  autoregression; the real $\tau_b$ sits well below it. ``Tok/call'' is the mean content
  tokens committed per accept-call; ``Same-call'' is the fraction of token pairs
  committing in one call (ties). A block-$\tau_b$ sweep over bin size is in
  Figure~\ref{fig:order} (right).}
  \label{tab:order}
  \begin{tabular}{llrrrr}
    \toprule
    Regime & Task & Token $\tau_b$ [95\% CI] & Block-seq.\ control & Tok/call & Same-call \\
    \midrule
    R1 & GSM8K (math)        & $0.512$ $[0.456,0.569]$   & $0.96$ & $25.5$ & $0.28$ \\ 
    R2 & HumanEval (code)    & $0.430$ $[0.318,0.527]$   & $0.94$ & $23.0$ & $0.49$ \\ 
    R3 & MBPP (code)         & $0.604$ $[0.561,0.647]$   & $0.96$ & $19.8$ & $0.19$ \\ 
    R4 & Factual             & $0.460$ $[0.422,0.496]$   & $0.41$ & $5.8$  & $0.50$ \\ 
    R5 & Open-ended          & $0.502$ $[0.463,0.542]$   & $0.96$ & $12.6$ & $0.11$ \\ 
    R6 & JSON (constrained)  & $-0.044$ $[-0.086,-0.00]$ & $0.75$ & $8.5$  & $0.72$ \\ 
    \bottomrule
  \end{tabular}
\end{table}

\paragraph{Regime dependence: JSON is approximately order-independent.} The structured
JSON regime (R6) is distinct: its token-level $\tau_b{=}-0.044$ with prompt-clustered
95\% CI $[-0.086,-0.00]$.
\ With its upper bound essentially at zero, JSON is \emph{approximately
order-independent} rather than strongly non-sequential, in contrast to the moderate
positive bias of the other regimes. This regime dependence is consistent with prior
reports of structural, out-of-order anchoring in constrained
generation~\citep{2602.02112v2,2601.15593v2}.

\paragraph{Robustness to commit non-monotonicity.} Commits are not strictly frozen: over
the 600 generations we observe 4524 un-accept events (mean $7.5$/gen), and only $220/600$
generations are fully monotone (\S\ref{sec:method}).
\ Re-computing $\tau_b$ on the strictly monotone subset gives values close to the full
sample (R1 $0.579$ vs $0.512$; R2 $0.386$ vs $0.430$; R4 $0.458$ vs $0.460$; R6 $-0.019$
vs $-0.044$),
\ so the un-accept behaviour does not bias the order conclusion. The non-monotonicity is
itself a model property worth further study (\S\ref{sec:threats}).

\subsection{Confidence and correctness, and early stopping}
\label{sec:results-calib}

\begin{figure}[t]
\centering
\includegraphics[width=0.62\linewidth]{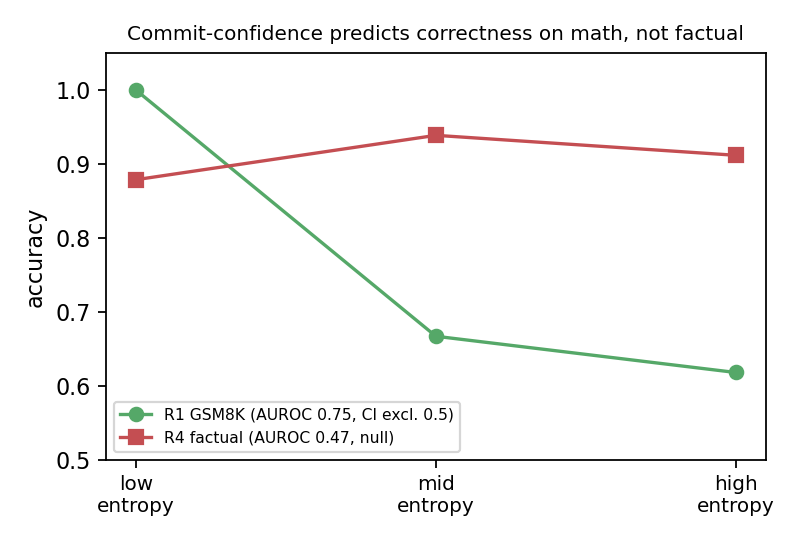}
\caption{Within-regime reliability of commit confidence by entropy tertile (low/medium/
high). On GSM8K (math, R1) accuracy falls monotonically as commit entropy rises
($1.00/0.667/0.618$; AUROC $0.749$, clustered CI excluding chance), so confidence predicts
correctness. On factual recall (R4) the curve is flat ($0.879/0.939/0.912$; AUROC $0.471$,
CI includes $0.5$), a null. Commit-confidence is predictive on math but not factual.}
\label{fig:calib}
\end{figure}

Table~\ref{tab:calib} summarizes how well within-regime commit confidence (negative
entropy-at-commit) discriminates correct from incorrect generations, now with
prompt-clustered AUROC CIs and a reliability-by-entropy-tertile curve. The result is
\emph{regime-specific}. On GSM8K (math), accuracy is $0.76$ and the AUROC of
$-\text{entropy} \rightarrow \text{correct}$ is $0.749$, prompt-clustered 95\% CI
$[0.602,0.879]$, which excludes chance; reliability by entropy tertile
(low$\to$high) is monotone, $1.00 / 0.667 / 0.618$ (24 errors).
\ So on math, lower commit entropy genuinely predicts correctness. On factual recall the
AUROC is $0.471$, CI $[0.383,0.544]$, which includes $0.5$ (a null result), with
flat tertiles ($0.879 / 0.939 / 0.912$; 9 errors).
\ The JSON regime has $1.000$ accuracy with no errors, so AUROC is undefined there.
\ Commit confidence therefore predicts correctness on math but not on factual recall
(Figure~\ref{fig:calib}); the signal is regime-specific and informative only where errors
occur (\S\ref{sec:threats}).

\begin{table}[t]
  \centering
  \caption{Within-regime discrimination of commit confidence (negative entropy) against
  correctness on the held-out strengthening run (5 seeds $\times$ 20 prompts). AUROC is
  for $-\text{entropy} \rightarrow \text{correct}$ with a prompt-clustered 95\% bootstrap
  CI. The last three columns are accuracy by entropy tertile (low/medium/high), a
  reliability curve. Math is real (CI excludes chance, monotone tertiles); factual is null
  (CI includes $0.5$, flat tertiles). Pooling across regimes is invalid (see text).}
  \label{tab:calib}
  \begin{tabular}{llrlrrr}
    \toprule
    Regime & Task & Acc & AUROC [95\% CI] & Tert.\ low & mid & high \\
    \midrule
    R1 & GSM8K (math) & $0.76$ & $0.749$ $[0.602,0.879]$ & $1.00$  & $0.667$ & $0.618$ \\ 
    R4 & Factual      & $0.91$ & $0.471$ $[0.383,0.544]$ & $0.879$ & $0.939$ & $0.912$ \\ 
    R6 & JSON         & $1.00$ & n/a (no errors)         & ---     & ---     & ---     \\ 
    \bottomrule
  \end{tabular}
\end{table}

\paragraph{A pooling caution, not a finding.} Pooling commitments across regimes into a
single AUROC gives $0.437$,
\ numerically below $0.5$, which might be misread as ``more confident
$\rightarrow$ more often wrong.'' This is a Simpson's-paradox artifact: regimes differ
in both base accuracy and entropy scale, so the pooled correlation can reverse the
within-regime trend. We report it as a methodological caution, not as evidence of
miscalibration; the valid statements are the within-regime ones above.

\paragraph{Aggressive entropy-bounded early stopping.} Despite a 48-step budget,
generations finish in only 3.3--17.1 accept-calls on average
\ (R4 factual $3.3$; R5 open-ended $17.1$), tracking content length. The
entropy-at-commit is $0.002$--$0.012$ nats,
\ one to two orders of magnitude below the $0.1$ entropy bound,
\ so commitments fire well inside the confidence threshold. In effect the model
commits content in a small late burst of accept-calls rather than using its full
denoising budget.

\subsection{Accuracy contrast with the autoregressive sibling}
\label{sec:results-ar}

Table~\ref{tab:ar} contrasts DiffusionGemma's accuracy with the matched AR sibling
\armodelid{} on the three scorable regimes. The AR model scores
$0.70$ (R1), $0.967$ (R4) and $1.00$ (R6);
\ DiffusionGemma scores $0.66$--$0.73$ (R1), $0.93$--$0.96$ (R4) and $1.00$ (R6).
\ Accuracy is thus comparable across the scorable regimes. We did not power this
comparison for an equivalence test ($n{\approx}30$ prompts per cell, no
pre-specified equivalence bound, no CIs on the differences), so it cannot rule out
moderate differences in either direction (\S\ref{sec:threats}). We draw no latency
conclusion beyond the step-count observations in \S\ref{sec:results-calib}, and the AR
model's commit order is left-to-right by construction, so it is not a fair comparand for
DiffusionGemma's measured $\tau_b$.

\begin{table}[t]
  \centering
  \caption{Accuracy contrast against the matched autoregressive sibling Gemma-4
  26B-A4B on the scorable regimes. DiffusionGemma's R1 range spans the exploratory
  deep run and the confirmatory run. Accuracy is comparable; the comparison
  ($n{\approx}30$ prompts per regime, no CI, no equivalence bound) was not powered for an
  equivalence test.}
  \label{tab:ar}
  \begin{tabular}{lrr}
    \toprule
    Regime & AR Gemma-4 26B-A4B & DiffusionGemma \\
    \midrule
    R1 (GSM8K)  & $0.70$  & $0.66$--$0.73$ \\ 
    R4 (Factual) & $0.967$ & $0.93$--$0.96$ \\ 
    R6 (JSON)   & $1.00$  & $1.00$ \\ 
    \bottomrule
  \end{tabular}
\end{table}

\paragraph{Findings beyond the model card.} The study meets its pre-registered success
bar (a target fixed before analysis: at least three quantified findings absent from
Google's public blog and model card)
\ via the partial granularity-dependent commit-order structure, regime-dependent
(approximately order-independent) JSON ordering, regime-specific
confidence--correctness discrimination (math real, factual null), and aggressive early
stopping, and answers the primary order hypothesis with prompt-clustered $\tau_b \pm$ CI
across all six regimes.

\section{Discussion and Limitations}

The picture that emerges is more careful than our earlier reading. DiffusionGemma has a
\emph{partial, granularity-dependent} left-to-right commit bias, not clean
block-autoregression: token order is only moderate, block-$\tau_b$ rises smoothly with
bin size (no privileged block at 16), the real order sits well below a block-sequential
control (genuine sub-block disorder), and large commit batches leave within-batch order
unresolved. Structured JSON output is approximately order-independent rather than strongly
non-sequential. Its shipped entropy-bounded sampler commits early, using a small fraction
of the 48-step budget, and its commit confidence predicts correctness on math but is null
on factual recall: a regime-specific, not universal, signal. On the scorable regimes its
accuracy is comparable to the AR sibling. The
result is a tempered version of the split in prior work between
``secretly left-to-right''~\citep{2512.21446v2,2506.00413v2} and ``structured out-of-order''~\citep{2602.02112v2,2601.15593v2} accounts: the model leans left-to-right
only partially, more so at coarser granularity, and only in some regimes.

\subsection{Threats to validity}
\label{sec:threats}

\paragraph{Construct validity of $\tau_b$ as ``order''.} Kendall $\tau_b$ between commit
call and position measures a rank correlation, not a mechanism. A high $\tau_b$ is
consistent with a left-to-right commit bias but does not prove the model internally
reasons left-to-right; we report $\tau_b$ as a descriptive order statistic and avoid
causal language.

\paragraph{Within-batch order is unresolved (coarse commit resolution).} Commit is
observed at accept-call resolution, and commit batches are large: $\approx 13$--$26$
content tokens commit per accept-call (E38),
\ with up to $0.72$ of token pairs committing in the same call (E39).
\ Within-batch order is therefore unresolved, and pairs that share a call are
simultaneous rather than out of order; our reported out-of-order pairs include these
same-call ties. We use the tie-aware Kendall $\tau_b$ for this reason, but the
token-level interpretation should be read as an accept-call-level tendency, not a strict
within-step ordering. This is the main reason the token-$\tau_b$ sits below the
block-sequential control: much of the sub-block ``disorder'' is genuinely unresolved
order, not measured reordering.

\paragraph{Commits are not strictly frozen (non-monotonicity).} We record the
\emph{first} accept-call that commits a position. Commits are \emph{not} strictly
monotone: over 600 generations we observe 4524 un-accept events (a committed position
returns to masked; mean $7.5$/gen) and only $220/600$ generations are fully monotone
(E42).
\ ``Commit order'' is therefore first-acceptance order. Reassuringly, $\tau_b$ on the
strictly monotone subset is close to the full-sample value (E43),
\ so the order conclusion is robust to this re-masking. The un-accept behaviour is itself
a model property worth further study.

\paragraph{Block size is an analysis choice (sweep shown).} The 16-token bin is an
analyst-chosen analysis granularity, \emph{not} the model's architectural block size.
Rather than rely on a single bin, we sweep $B \in \{4,8,16,32,64\}$ and find block-$\tau_b$
rises \emph{smoothly} with $B$ (E37),
\ with no jump at 16; coarser bins look more sequential because they hide more within-bin
disorder. We also bound clean block autoregression with a block-sequential control (E41),
\ which the real order falls well below. The remaining limitation is that any single
bin-level number is granularity-dependent and should be read off the sweep, not in
isolation.

\paragraph{Unit of analysis, seeds, and bootstrap clustering.} The headline statistics
come from 5 seeds $\times$ 20 prompts per regime, and the bootstrap is
\emph{prompt-clustered} (resampling the 20 prompts with their seeds), so the CIs respect
within-prompt cross-seed correlation. Seed variance is small (SD of mean-$\tau$
$0.014$--$0.050$, E44),
\ so the conclusions are stable across seeds. The remaining limitation is the modest
prompt count (20/regime) and a single checkpoint; broader replication would further
tighten the intervals.

\paragraph{Calibration is informative only where errors occur; regime-specific.} The
confidence--correctness AUROCs now carry prompt-clustered CIs and a reliability-by-tertile
curve, and the result is \emph{regime-specific}: real on math (CI excludes chance, monotone
tertiles, E45)
\ but null on factual recall (CI includes $0.5$, flat tertiles, E46);
\ the earlier strong factual AUROC was small-$n$ noise. The signal is meaningful only on
regimes with a spread of correct and incorrect outcomes, and even there we report AUROC and
a tertile reliability curve but no full ECE. R6 has no errors (E47),
\ so its AUROC is undefined and we make no claim there.

\paragraph{Multiplicity.} We report intervals across six regimes and multiple statistics
without multiplicity correction; we read the CI statements (e.g.\ the GSM8K AUROC CI
excluding chance, or the JSON $\tau_b$) descriptively, not as multiplicity-controlled
significance tests.

\paragraph{Out-of-order rate is a restatement of $\tau_b$.} The confirmatory out-of-order
rate ($0.086$--$0.208$, E40)
\ is a monotone restatement of $\tau_b$ rather than independent corroboration of the order
finding.

\paragraph{Trailing-EOS padding artifact (caught and corrected).} On raw canvases the
naive order metric gives $\tau_b \approx -0.9$,
\ because trailing EOS/pad positions are committed first and dominate the correlation.
Restricting to content positions (\S\ref{sec:method}) removes this artifact; all
reported numbers are content-only. This is a concrete pitfall for any commit-order
study on padded canvases.

\paragraph{Simpson's-paradox pooling (caught and corrected).} As shown in
\S\ref{sec:results-calib}, pooling the confidence--correctness analysis across regimes
yields an AUROC of $0.437$
\ (numerically below $0.5$; we do not claim it differs from chance) that reverses the
valid within-regime trend. We treat this only as a caution and report the analysis
strictly within regimes.

\paragraph{Unscored code regimes.} R2/R3 have no code-execution harness, so we report
no correctness for them; they inform the order analysis only. This limits the
confidence--correctness claim to math and factual recall.

\paragraph{AR contrast is on accuracy only, and underpowered.} The AR sibling commits
left-to-right by construction; its ``order'' is not measured and is not a fair $\tau_b$
comparand, so the contrast is on accuracy only. Accuracy is comparable across the
scorable regimes, but we did not power this comparison for an equivalence test (no
equivalence bound, no CIs on the differences, $n{\approx}30$ per cell), so it cannot
establish parity.

\paragraph{Single GPU, single checkpoint, small constrained regimes.} All runs are on one
H100; the strengthening run uses 5 seeds but 20 prompts per regime, and R4--R6 are small.
Confidence intervals reflect this, but broader replication (larger constrained sets,
additional checkpoints) would strengthen the regime-dependence and confidence--correctness
claims. Total compute was approximately 0.9 H100-hours.
\ R1--R3 may also be subject to pretraining contamination, which would bias accuracy
reads but not the decoding-order spine (order is \emph{how} the model generates,
independent of memorization).

\section{Availability}

No code, data, or checkpoints associated with this study are publicly released. We do
not maintain a public repository for these artifacts. The instrumentation harness, the
probe suite, and the recorded traces are available from the authors on request.

\section{Conclusion}

We measured how a public diffusion language model actually decodes. By hooking
DiffusionGemma's own commit mechanism, we found a \emph{partial, granularity-dependent}
left-to-right commit bias, moderate at the token level and rising smoothly toward
sequential at coarser analysis bins, with no privileged block at 16, rather than clean
block-autoregression: the real order sits well below a block-sequential control, and large
commit batches leave within-batch order unresolved. The bias is regime-dependent, with
structured JSON output approximately order-independent. Its entropy-bounded sampler commits
early and well inside its confidence threshold; its commit confidence predicts correctness
on math but not on factual recall (a regime-specific signal, with a clustered AUROC CI and a
reliability curve); and its accuracy is comparable to the matched autoregressive
sibling on the scorable regimes. A strengthening run (5 seeds,
prompt-clustered bootstrap, a block-size sweep, a block-sequential control, and a
commit-non-monotonicity check) walked back our initial ``block-autoregressive'' headline to
this more careful claim. Along the way we documented measurement artifacts and checks
(EOS padding, Simpson's-paradox pooling, commit non-monotonicity, block-size sensitivity,
and commit-batch ties) that we expect to recur in commit-order studies. We hope the
instrumentation lens, observing a shipped sampler's per-step commitments rather than
proposing a new one, proves useful for auditing other diffusion language models.

\bibliographystyle{plainnat}
\bibliography{references}

\end{document}